# The Locally Deployable Virtual Doctor: LLM Based Human Interface for Automated Anamnesis and Database Conversion

Jan Benedikt Ruhland[1], Doguhan Bahcivan[1], Jan-Peter Sowa[2], Ali Canbay[2], Dominik Heider[3,4,]*

1: Heinrich Heine University Düsseldorf, Faculty of Mathematics and Natural Sciences, Düsseldorf, Germany
2: University Hospital Knappschaftskrankenhaus Bochum, Department of Internal Medicine, Bochum, Germany
3: Philipps University of Marburg, Department of Mathematics and Computer Science, Marburg, Germany
4: University of Münster, Institute of Medical Informatics, Münster, Germany

*corresponding author: dominik.heider@uni-muenster.de

## Abstract

Recent advances in large language models made it possible to achieve high conversational performance with substantially reduced computational demands, enabling practical on-site deployment in clinical environments. Such progress allows for local integration of AI systems that uphold strict data protection and patient privacy requirements, yet their secure implementation in medicine necessitates careful consideration of ethical, regulatory, and technical constraints.

In this study, we introduce MedChat, a locally deployable virtual physician framework that integrates an LLM-based medical chatbot with a diffusion-driven avatar for automated and structured anamnesis. The chatbot was fine-tuned using a hybrid corpus of real and synthetically generated medical dialogues, while model efficiency was optimized via Low-Rank Adaptation. A secure and isolated database interface was implemented to ensure complete separation between patient data and the model's inference process. The avatar component was realized through a conditional diffusion model operating in latent space, trained on researcher video datasets and synchronized with mel-frequency audio features for realistic speech and facial animation.

Unlike existing cloud-based systems, this work demonstrates the feasibility of a fully offline, locally deployable LLM-diffusion framework for clinical anamnesis. The autoencoder and diffusion networks exhibited smooth convergence (final L1 = 0.037, MSE = 0.016), and MedChat achieved stable fine-tuning with strong generalization to unseen data. The proposed system thus provides a privacy-preserving, resource-efficient foundation for AI-assisted clinical anamnesis, also in low-cost settings.

Keywords: Machine Learning, AI in medicine, LLM, Diffusion Model, Autoencoder

# 1. Introduction

Artificial intelligence (AI) has reached a pivotal moment in its development, leading to transformative changes across nearly all sectors of human society [1-4]. The rapid evolution of AI technologies presents both significant opportunities and substantial challenges, particularly with respect to adaptation, regulation, and ethical implementation. Central to this advancement is the emergence and continuous improvement of Large Language Models (LLMs), which are not only driving innovation in natural language processing and human-computer interaction but also in other fields [2, 5].

One prominent example is ChatGPT [6], an easily accessible and highly versatile LLM that has demonstrated utility across a wide range of applications. These include grammar correction, content refinement, application development, and even educational support. Despite their flexibility and impressive capabilities, such models possess significant risks. They may produce incorrect or misleading outputs, especially when their responses are accepted without critical scrutiny [7]. Nevertheless, the societal potential of LLMs remains vast, particularly when appropriately integrated into domain-specific workflows.

In the field of medicine, AI occupies a uniquely sensitive and impactful position. The application of machine learning to medical data requires rigorous considerations due to the inherently private, complex, and ethically sensitive nature of healthcare information [8, 9]. Ensuring data privacy, maintaining transparency, and adhering to ethical research practices are essential prerequisites that complicate the implementation of AI in clinical settings. However, despite these challenges, AI has already demonstrated its capacity to augment human expertise. For instance, convolutional neural networks have been reported to show superior performance in certain diagnostic tasks, particularly in medical imaging and pattern recognition [10, 11].

One area where AI has the potential to further improve healthcare delivery is in anamnesis, the medical history-taking process. Effective anamnesis is complicated by the multivariate nature of symptoms, disease-specific manifestations, and the potential for asymptomatic progression, as seen in conditions such as COVID-19 [12]. Traditional approaches often rely on standardized questionnaires, which may overlook critical individual nuances. In contrast, LLMs offer the flexibility to adapt questions dynamically based on patient responses, allowing for a more personalized and comprehensive data collection process.

By integrating LLMs into anamnesis workflows, clinicians could receive more structured, relevant, and individualized information, potentially leading to earlier and more accurate diagnoses. Much like the role of CNNs in medical image tasks, LLMs could serve as decision-support tools that assist healthcare professionals, reduce the risk of diagnostic errors, and contribute to the early identification of severe diseases. Thus, with appropriate safeguards, AI holds promise as a transformative aid in the practice of modern medicine.

As demonstrated in a recent study by Tu et al., LLMs have begun to outperform medical professionals in real-world clinical settings [13]. Their model, the Articulate Medical Intelligence Explorer (AMIE), was evaluated in a controlled study where trained actors simulated patient interactions, allowing for direct comparison between AMIE and practicing clinicians during the anamnesis process. The objective of the study was to assess the viability of deploying a domain-specific LLM for clinical use, particularly in patient history-taking. The results revealed that AMIE, even when used by non-experts, outperformed licensed medical professionals across multiple diagnostic dimensions. This underscores the transformative potential of LLMs in supporting or even enhancing the clinical decision-making process, especially in initial diagnostic stages such as anamnesis.

Our study aimed to develop a locally deployable LLM, named MedChat, optimized for use in real-world clinical environments with constrained hardware capabilities. A key objective was to ensure secure integration with patient data systems. To that end, MedChat was designed to access a patient's medical history only through a pre-processing interface: the relevant data is extracted and securely fed to the model prior to the interaction, without granting the LLM direct access to the underlying database. This approach balances the benefits of personalized context with stringent privacy and security standards. In addition, safeguards were implemented to mitigate risks associated with misuse or unauthorized manipulation of the chatbot, addressing a common concern in deploying AI systems in sensitive domains such as healthcare.

Beyond the development of the LLM itself, we also designed a human-machine interface to enhance the interaction quality and user engagement. To achieve this, video recordings of two researchers were used to train a diffusion model operating on latent space transformations, capable of generating synchronized speech and facial animations based on textual input. This enabled the creation of a virtual physician avatar, effectively simulating a human-like presence without requiring real-time human supervision. The resulting system offers a fully automated, responsive, and lifelike clinical interaction environment, potentially serving as a scalable solution for digital anamnesis.

The main contributions of this work are summarized as follows:

1. We present MedChat, a locally deployable large language model tailored for clinical anamnesis, capable of operating fully offline on consumer-grade hardware.

2. We integrate a diffusion-based virtual physician avatar for naturalistic, multimodal interaction combining speech and facial animation.

3. We employ parameter-efficient fine-tuning (LoRA) to achieve medical adaptation while maintaining the general-purpose capabilities of the base model.

4. We propose a secure database interface that enables structured, privacy-preserving integration of patient information without granting direct database access to the language model.

5. We demonstrate the complete end-to-end training and evaluation of all system components, laying the groundwork for future clinical validation.

## Research in Context

The integration of LLMs into healthcare has advanced rapidly, encompassing diverse applications ranging from diagnostic assistance and clinical documentation support to systems that approach partial or near-full autonomy in tasks traditionally performed by physicians, such as patient consultation and treatment planning [13, 14]. Current implementations include clinical decision support systems [14], symptom-based triage agents [15], and automated virtual health advisors [16]. More experimental research has focused on developing fully autonomous "Virtual Doctor" systems capable of interacting directly with patients; however, these remain largely at the conceptual or prototype stage due to persistent ethical, regulatory, and safety challenges [17].

A major limitation in the deployment of LLMs and their embodied counterparts lies in their substantial computational requirements. High-performance models demand extensive memory resources and GPU processing capacity [18], often making cloud-based deployment necessary. While such infrastructure alleviates the need for on-premise hardware, it introduces significant concerns related to privacy, data protection, and compliance with medical data governance frameworks, particularly when sensitive information such as patient records, imaging data, or personal health histories is processed.

Most existing LLM-based healthcare systems rely on purely text-based chatbot interfaces that lack visual embodiment or avatar representation [13-16]. Although these models provide robust functional capabilities, their absence of visual and emotional expressiveness can reduce patient engagement and trust, elements that are fundamental in clinical communication and adherence. To address this limitation, recent research has investigated the integration of avatars and virtual agents capable of visual and nonverbal interaction [19, 20].

Unlike avatars developed through game-engine-based real-time rendering or pre-recorded video playback, AI-driven avatar systems are typically implemented using generative approaches such as Generative Adversarial Networks [21] or diffusion-based models [22]. These architectures enable the synthesis of realistic facial expressions, gestures, and lip-synchronized speech, offering the potential for more natural and immersive interactions. However, achieving real-time or near-real-time responsiveness remains a considerable technical challenge. Full-scale diffusion models exhibit high latency, making them unsuitable for dynamic, interactive clinical environments. Consequently, researchers have explored lightweight alternatives, including optimized GANs and parameter-efficient diffusion models, often referred to as "fast" or "turbo" diffusion variants, that aim to balance visual fidelity with computational efficiency [23, 24].

Despite these advancements, avatar-based or visually embodied LLM systems continue to face substantial computational barriers. These architectures require not only sufficient GPU memory to host the LLM and associated generative modules, often exceeding 12-16 GB for high-resolution outputs, but also high-throughput tensor processing to meet the latency constraints of real-time interaction. Accordingly, the hardware and performance demands of such systems remain a significant obstacle to their large-scale deployment and integration within clinical workflows.

In contrast to prior work, our approach uniquely combines local deployability, multimodal avatar-based communication, and strict data isolation mechanisms. Whereas systems such as AMIE or MedAssist rely on cloud infrastructure with extensive computational demands, MedChat demonstrates that comparable performance can be achieved under resource constraints while maintaining full data sovereignty and GDPR (General Data Protection Regulation) compliance. Table 1 provides a comparison of our model MedChat and other existing systems.

**Table 1:** Comparison of existing LLM-based medical systems.

| System | Deployment | Avatar Interface | Privacy Model | Clinical Readiness |
|---|---|---|---|---|
| AMIE [13] | Cloud | No | Limited | Experimental |
| MedAssist [14] | Cloud | No | Partial | Prototype |
| HealsHealthAI [16] | Cloud | No | Partial | Research |
| MedChat | Local | Yes | Full Isolation | Preclinical Evaluation Planned |

# 2. Materials and Methods

## 2.1 Project Overview

The primary objective of this project was to design and implement a locally deployable virtual physician system, referred to as MedChat, capable of operating entirely offline within clinical environments. The system is specifically developed to function without reliance on external servers or internet connectivity, thereby ensuring compliance with strict data protection regulations and maintaining the highest possible standards of patient privacy and information security.

The MedChat system is conceptualized as an interactive virtual physician embedded within a controlled physical environment. In a typical clinical workflow, the patient enters a private, designated consultation room equipped with a high-resolution display, camera, and computing interface. Within this space, the virtual physician is projected on the screen, enabling a naturalistic and confidential interaction. Prior to initiating the consultation, the patient is registered by clinical staff, and relevant portions of the patient's medical history are securely preloaded into the system's local database. This allows the virtual physician to access contextually relevant information during the anamnesis process while preserving data locality and avoiding external data transmission.

During the consultation, MedChat autonomously conducts a structured anamnesis, guiding the patient through a series of context-aware questions designed to capture essential clinical information. The collected data are then processed, organized, and securely stored in a local database using encrypted storage protocols. Following each session, the system automatically generates a structured summary report that synthesizes the key findings of the interaction. This report can subsequently be accessed and reviewed by licensed healthcare professionals, providing an efficient and standardized foundation for further clinical assessment and decision-making.

Figure 1 presents a schematic overview of the anamnesis workflow facilitated by the MedChat system. By partially automating the patient intake procedure, MedChat contributes to the standardization of anamnesis across healthcare facilities, enhances the consistency of data collection, and optimizes the utilization of patient waiting times. Moreover, by reducing the administrative burden on medical personnel, the system allows healthcare providers to allocate more time to direct patient care and complex diagnostic reasoning. In this way, MedChat serves as both a technological innovation and a workflow enhancement tool, bridging artificial intelligence capabilities with the practical requirements of clinical practice.

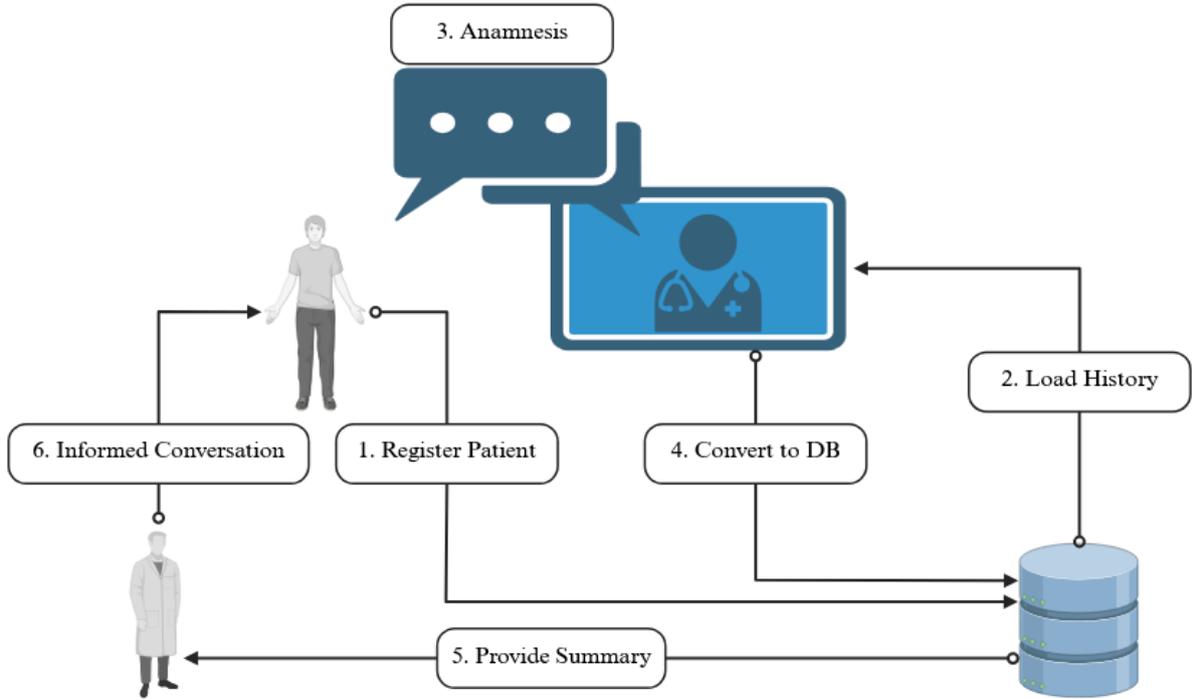

**Figure 1**: Overview of the anamnesis process facilitated by the MedChat system, illustrating its integration within standard clinical workflows.

## 2.2 The Virtual Doctor Avatar

### 2.2.1. Network Design and Preprocessing

We developed a comprehensive preprocessing pipeline to extract both audio and video frames from recorded materials. The audio signals were resampled to 22,050 Hz to enhance the fine-tuning performance of the text-to-speech (TTS) model. Prior to transformation, the recordings were downsampled to 16 kHz to ensure compatibility with the Wav2Vec model [25], which was employed for automated transcription. Pretrained Wav2Vec models from [26] were utilized to generate bilingual transcriptions in English and German. Subsequently, the audio data were converted into mel-frequency spectrograms using a window size of 1,024 and a hop length of 128, extracting 80 mel-frequency components to capture perceptually salient acoustic features.

The speech synthesis component was implemented using the Piper TTS framework [27], which follows an adversarial variational design inspired by the HiFi-GAN [28] and VITS [29] architectures. The training objective combines several complementary loss terms: (i) an adversarial loss applied to a least-squares discriminator-generator pair, (ii) a feature (perceptual) loss that minimizes the L1 distance between intermediate feature maps of real and generated audio, thereby enhancing perceptual fidelity in the frequency domain, and (iii) a Kullback-Leibler (KL) divergence term that regularizes the latent variable distribution. Together, these losses encourage both realistic waveform synthesis and consistent latent space structure, yielding speech output that closely resembles the target speaker's tone and prosody. For this purpose, we adapted the pretrained Thorsten Voice model [30] for German and the Amy model [31] for English, training both on voice recordings provided by two researchers. The resulting customized voices ensured natural and consistent speech output in both languages, closely resembling the speaker's tone and prosody.

To efficiently encode visual information, we developed a custom autoencoder that compresses image data into a compact latent-space representation, thereby reducing the computational cost of subsequent video generation. The model architecture was inspired by the Stable Diffusion framework [32] and trained on facial image datasets supplied by the two researchers with video recordings around 30 minutes yielding over 70,000 images each with corresponding audio signals. For initial experiments, we additionally employed the "Visual Historical Record of American High School Yearbooks" dataset [33] to assess model generalization. The latent space was constrained to the range [-1, 1] using a hyperbolic tangent activation function to enhance numerical stability. Although this constraint may distort direct decoding, it improves robustness to noise and supports high-quality image reconstruction. To mitigate potential distortion, an adaptive normalization layer [34] was integrated into the decoder to map the constrained latent space back into an unconstrained representation. Furthermore, noise augmentation was introduced into the latent space to simulate imperfections typical of diffusion processes, promoting resilience during image generation. The autoencoder was designed to operate independently of input resolution, maintaining proportional compression relative to the image dimensions. Details of the model architecture and experimental results supporting these design choices are provided in appendix A.

For the video generation stage, a conditional diffusion model was implemented, conditioned jointly on the input image and the mel-frequency representation of the corresponding audio frame. This multimodal conditioning enables synchronized lip movement and facial dynamics consistent with the spoken content. The diffusion model adopts a U-Net backbone architecture augmented with cross-attention layers, conditional embeddings, and multi-head self-attention mechanisms, ensuring both temporal coherence across sequential frames and semantic consistency between visual and auditory modalities.

During inference, the model diverges from conventional diffusion frameworks that typically initiate the generation process from random Gaussian noise. Instead, it begins from a pre-diffused default image, which serves as a structured prior, substantially reducing the number of denoising steps required for image reconstruction. This modification results in a significant decrease in computational cost and inference latency while maintaining high visual fidelity in the synthesized video output.

Furthermore, the system was evaluated for its ability to generate multiple subsequent frames, from the same pre-diffused input. This approach enables parallel frame generation, potentially bridging the performance gap between diffusion-based synthesis and real-time video rendering. The resulting architecture demonstrates that, with appropriate conditioning and initialization strategies, diffusion models can achieve interactive responsiveness suitable for real-time avatar-based communication systems.

Figure 2 illustrates the training process of the diffusion model, highlighting the key modifications introduced during inference. During training, the model receives as input the current frame, the mel-frequency representation of the corresponding audio, and the previous frame, in addition to the timestep provided by the cosine noise scheduler. The target frame is progressively diffused, and the network is trained to predict the added noise, enabling it to learn the denoising dynamics. During inference, the trained model generates a sequence of frames, corresponding to one second of video (30 frames), by conditioning on a fixed reference image and a weakly diffused version of the same image. The mel-frequency inputs are updated continuously to reflect the evolving audio, allowing the system to produce temporally coherent video frames that are synchronized with the speech signal. This approach ensures both high-fidelity reconstruction and real-time responsiveness in the Virtual Doctor Avatar.

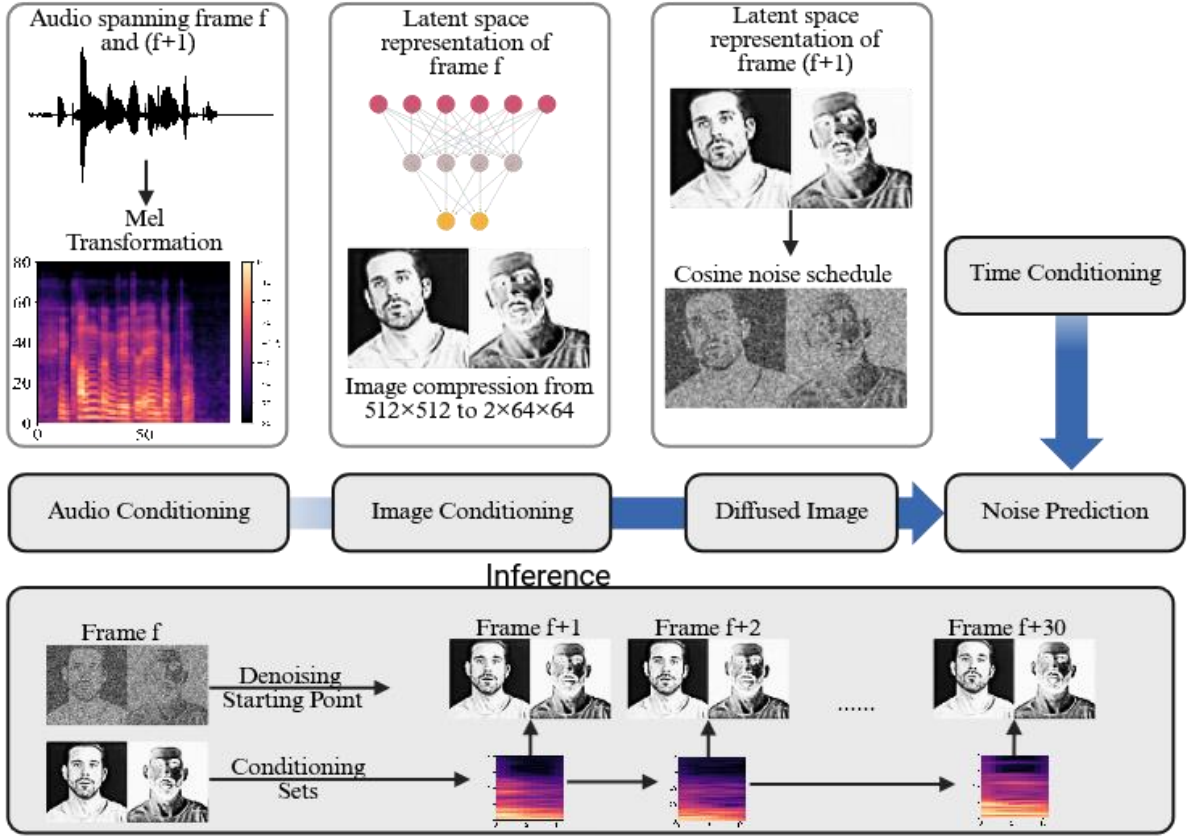

**Figure 2**: Illustration of the diffusion model training and inference pipeline for the Virtual Doctor Avatar. During training, the model is conditioned on mel-frequency features, the previous frame, and the current timestep from the cosine noise scheduler. The target frame is progressively diffused and used to train the network to predict the added noise. During inference, the trained model generates one second of video (30 frames) using a consistent conditioning image and a weakly diffused version of that image, while the conditioning mel-frequency inputs vary over time to reflect the changing audio features.

**Table 2:** Overview of model architectures and hyperparameters.

| Component | Architecture | Loss Function | Epochs | Learning Rate |
|---|---|---|---|---|
| Autoencoder | Encoder-Decoder CNN | L1 Reconstruction | 30 | 0.00001 |
| Diffusion U-Net | Conditional U-Net with cross-attention | MSE | 1,000 | 0.0001 |
| Piper TTS | Variational Inference GAN | Adversarial, L1, KL | 135 | 0.00001 |
| MedChat (LLM) | LLaMA 3.1-8B with LoRA | Cross-Entropy | 4 | 0.000005 |

### 2.2.2. Training Procedure

Having established the preprocessing and network design, we next detail the fine-tuning procedures applied to each component, focusing on training stability, convergence behavior, and computational efficiency. The training process commenced with the fine-tuning of the audio data using the Piper TTS model. A reduced learning rate of 0.00001 was applied to both the generator and discriminator components to promote stable convergence and prevent overfitting. Training was conducted over 135 epochs, with model checkpoints evaluated every 50 epochs to assess qualitative and quantitative improvements. After 135 epochs, the synthesized voice demonstrated a high degree of perceptual similarity to the target speaker. However, additional training beyond this point led to a gradual degradation in voice quality, suggesting the onset of overfitting and confirming the optimal stopping point for fine-tuning.

Following the completion of the speech synthesis stage, training proceeded to the video generation component, utilizing the previously developed autoencoder as the foundational model for latent space compression. After verifying the model's robustness on the public image dataset [33] (see appendix A), the network was fine-tuned on recordings of the target researcher for 30 epochs. Training employed the Adam optimizer with an initial learning rate of 0.00001 and an L1 reconstruction loss as the objective function. A batch size of 32 was used, with all input images standardized to a resolution of 512×512 pixels. The autoencoder's latent space was configured for a compression factor of 32.

For the diffusion-based generation stage, a cosine noise scheduling strategy [35] was employed to ensure smooth and stable convergence across 600 diffusion time steps. The model was trained for 1,000 epochs using the Adam optimizer with an initial learning rate of 0.0001, modulated by a cosine annealing schedule to facilitate adaptive step-size decay. The mean squared error (MSE) between the predicted and actual noise served as the training objective, consistent with standard Denoising Diffusion Probabilistic Model formulations [32].

## 2.3 MedChat Model Development

### 2.3.1 Datasets and Conversation Synthesis

To construct the primary dataset for large language model fine-tuning, we integrated three publicly available datasets from the Kaggle machine learning platform [36, 37, 38]. These datasets collectively comprised 10,080 samples, each containing a set of symptoms associated with a specific disease.

To generate realistic medical dialogues from these structured data samples, we employed a teacher-student knowledge distillation approach. Specifically, we fine-tuned the Meta-Llama-3.1-70B-Instruct model [39] to serve as a teacher model capable of generating high-quality, medically relevant conversations. The generated dialogues were then used to train a smaller, more deployment-friendly model, Meta-Llama-3.1-8B-Instruct, which forms the basis of our MedChat system.

We leveraged the capabilities of the larger, pretrained teacher model to reduce training time and benefit from its existing contextual understanding. However, due to the substantial hardware requirements (approximately 140 GB of VRAM in 16-bit precision) the 70B model is impractical for deployment in typical medical facility settings. Hence, our final MedChat implementation is based on the 8B version, which is more feasible for real-world applications.

To ensure the generated dialogues were clinically meaningful and structurally consistent, we provided the teacher model with detailed generation guidelines. These guidelines, described in the appendix B, included instructions on conversational tone, question structure, and appropriate responses to patient input. The model was allowed to generate up to 1,500 tokens per dialogue, enabling the creation of rich and informative medical conversations.

By repeating the generation process for every entry in the symptom-disease dataset, we created a large synthetic corpus of anamnesis dialogues. This synthetic dataset served as the foundation for fine-tuning our MedChat model, allowing it to effectively simulate medical interviews despite the absence of real patient interaction data.

Following the generation of 10,080 medical dialogues, we applied preprocessing steps to prepare the dialogues for model training, including formatting the conversations into structured JSON files. One common issue observed in the generated text was the excessive use of filler or acknowledgment phrases, e.g., "I see." or "Noted". To improve quality and coherence, we filtered out chatbot responses that did not contain a question and hence contained redundant information.

For training, 10% of the dataset was randomly withheld to serve as a validation set, enabling the assessment of overfitting and evaluation of model performance on an independent subset.

### 2.3.2 MedChat Training

We selected the Meta-Llama-3.1-8B-Instruct model [39] as the base model for fine-tuning, due to the impracticality of training a large language model from scratch, which would require extensive computational resources. Released in July 2024, this model is among the first robust large language models to combine a 128,000-token context window with multilingual capabilities, all within a manageable size of 8 billion parameters. Its relatively compact architecture allows it to be deployed and fine-tuned on consumer-grade or mid-range data center GPUs.

Fine-tuning was performed on a cluster of six NVIDIA A100 GPUs, each equipped with 40 GB of VRAM. We employed model parallelism by distributing instances of the model across all GPUs, splitting the dataset accordingly, and synchronizing gradients across devices during training. This setup enabled efficient parallel training and supported rapid experimentation. The dataset was divided into a 90:10 train-test split, with model evaluation conducted at the end of each epoch.

We used the AdamW optimizer [40] with a weight decay of 0.01 and an initial learning rate of 0.000005, coupled with a linear learning rate decay schedule. Each training batch consisted of 4 samples, and gradient accumulation was performed every 16 steps before synchronization across GPUs. Each epoch included 420 steps per GPU, and training was conducted for 4 epochs, resulting in approximately 1,680 training steps per device. In aggregate, this yielded 10,080 steps across all devices, corresponding to full coverage of our dataset.

To further enhance training efficiency and preserve the model's broad capabilities, we adopted a parameter-efficient fine-tuning approach using Low-Rank Adaptation (LoRA) [41]. Since the model is required to retain general-purpose functionalities, such as open-domain conversation, classification, summarization, and multilingual support, we aimed to avoid catastrophic forgetting. In line with the recommendations from [41], we limited fine-tuning to lightweight adapter modules inserted into the Key and Value projection matrices of the transformer layers. This selective update approach provides a favorable trade-off between computational cost and fine-tuning effectiveness while maintaining the integrity of the pretrained model's diverse skill set.

### 2.3.3 Database Integration

To manage and store data collected during interactions, we integrated an SQLite database into the MedChat system using SQLAlchemy, a Python-based Object-Relational Mapping framework. This architecture supports efficient and structured data handling, enabling seamless communication between the application and the underlying database layer.

Following the conclusion of each anamnesis session, MedChat produces a structured summary and classification of the dialogue in JSON format. This output is subsequently parsed, converted into application-level data objects, and persisted in the SQLite database. These operations (parsing, transformation, and data storage) are implemented outside the model itself and are managed by a dedicated service class, which functions as an intermediary between the model's output and the database. Although modern language models support direct database access via tool calls [39], we explicitly exclude this functionality to mitigate potential security risks and maintain strict control over data access.

To further safeguard the system against adversarial inputs, such as prompt injection and jailbreak attacks [42], we integrated Prompt Guard [43] into the application. Prompt Guard is a security-focused classification model that evaluates incoming prompts and categorizes them as benign, prompt injection, or jailbreak attempts. By embedding Prompt Guard into the input validation pipeline, we establish a proactive security layer that filters malicious content before it reaches the language model. This approach enhances the safety, reliability, and integrity of MedChat, particularly in the context of healthcare, where the consequences of model misuse can be significant.

# 3. Results

## 3.1 Training of the Autoencoder and Diffusion Model

The fine-tuning of the autoencoders, designed with a constrained latent space and augmented with latent-space noise, demonstrated stable and consistent convergence throughout the training process. The autoencoder was personalized using recordings from a single researcher, enabling the network to adapt to specific facial and vocal characteristics. Figure 3 presents the L1 reconstruction loss over the course of 30 training epochs. At the beginning of training, the L1 loss started at approximately 0.382 and rapidly decreased during the first 10 epochs, reaching a value of 0.051. The subsequent epochs exhibited a slower, steady decline, ultimately stabilizing around 0.037 by the final epoch. This smooth convergence indicates that the network successfully learned to encode and reconstruct input images while maintaining the constraints imposed on the latent space. The latent-space noise augmentation further contributed to this stability by preventing overfitting and promoting robustness to perturbations in the latent representation. Notably, the short fine-tuning duration of 30 epochs was sufficient to achieve a low reconstruction error, demonstrating the efficiency of the proposed architecture and training regimen.

The U-Net model employed for diffusion-based video generation also exhibited clear convergence over its training regime. Figure 4 illustrates the MSE value between the predicted and applied noise across 1,000 epochs. The training curve begins at an initial MSE of approximately 0.774 and shows a consistent downward trend, reaching around 0.016 by the final epochs. While minor fluctuations of up to ±0.002 are observed throughout training, these are attributable to the stochastic nature of the diffusion process, where noise is randomly sampled at each timestep, and do not indicate instability. The consistent overall decrease in MSE demonstrates that the network effectively learned to denoise latent representations while conditioning on both image and audio features. These fluctuations also suggest that the model maintains adaptability to varying noise patterns, which is beneficial for generalization during inference.

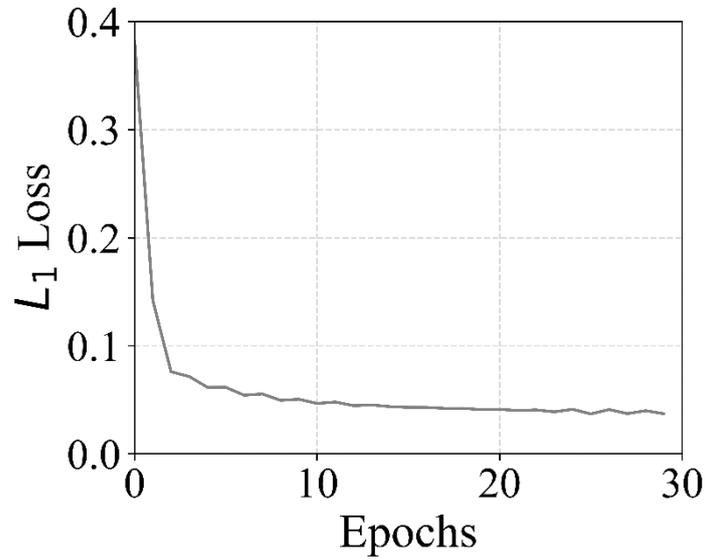

**Figure 3:** L1 reconstruction loss of the autoencoder during fine-tuning on recordings from a single researcher. The network was trained for 30 epochs with latent-space noise augmentation and a constrained latent representation. The curve demonstrates smooth and stable convergence, indicating effective learning of the input image reconstruction while maintaining latent-space regularity.

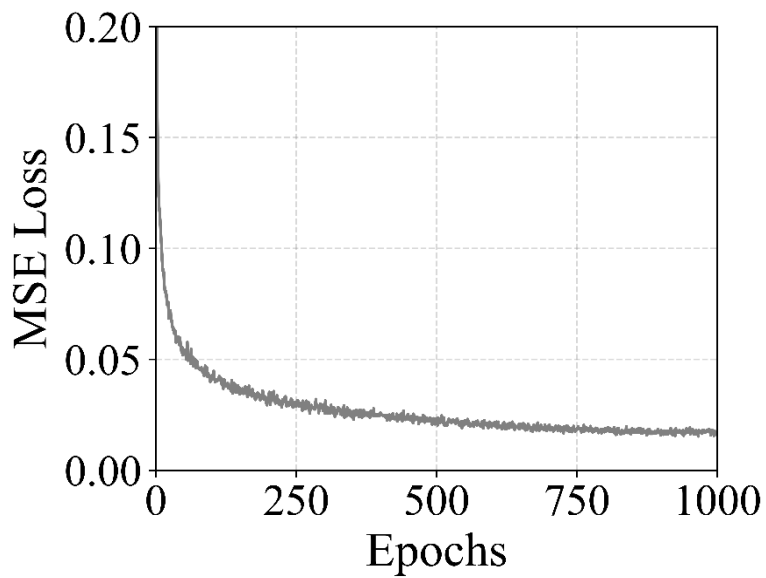

**Figure 4:** MSE value of the U-Net diffusion model during training over 1,000 epochs. The model predicts the noise added to latent representations conditioned on image and audio features using a cosine

noise schedule. Minor fluctuations are observed due to the stochastic nature of the diffusion process, while the overall downward trend demonstrates stable convergence.

## 3.2 Dialogue Generation and MedChat Fine-tuning

Figure 5 presents the fine-tuning dynamics of the MedChat model. At the onset of training, both the training and evaluation losses are relatively high, reflecting the model's initial lack of adaptation to the specific task and substantial deviation from the desired output distributions. During the first 500 training steps, a rapid decrease in both losses is observed, indicating that the model quickly begins to capture relevant patterns from the training data. Beyond this point, the training loss continues to decrease gradually, while the evaluation loss stabilizes, demonstrating that the model maintains consistent performance on unseen data. The close alignment between training and evaluation loss throughout the remainder of the fine-tuning process suggests minimal overfitting and highlights the model's ability to generalize effectively. Overall, these results indicate that the fine-tuning procedure successfully adapts MedChat to the target task while preserving robustness to novel inputs.

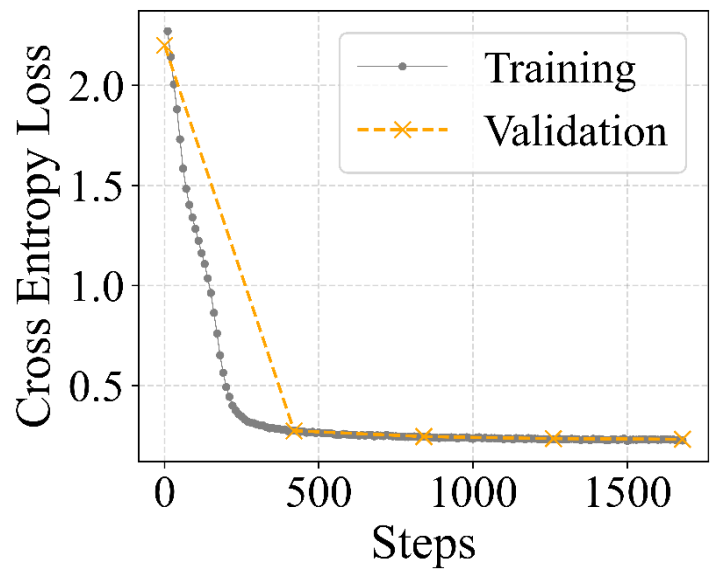

**Figure 5:** Training loss curves for the MedChat fine-tuning process. The model exhibits a rapid initial decrease in loss, followed by convergence, indicating effective learning and stable generalization to unseen data.

The simulated interaction described in Appendix C involves an initial consultation with a patient who has no prior history recorded in the MedChat system and for whom no external information from other models or sensor data is available. In this simulation, the patient is modeled as presenting symptoms consistent with pneumonia. The scenario is designed to evaluate the ability of MedChat to autonomously conduct a structured anamnesis, elicit relevant clinical information, and generate an initial patient summary entirely based on the interaction within the system. This setup provides a controlled environment for assessing the model's performance in handling cases without any pre-existing contextual knowledge or external guidance.

From a deployment perspective, the full MedChat-Avatar system can operate on consumer-grade GPUs with as little as 40 GB of VRAM for end-to-end interaction, or on 24 GB of VRAM and only CPUs when the audio-only mode is used. This configuration substantially reduces hardware costs and energy consumption compared to cloud-based LLM solutions, which typically require 80-100 GB VRAM or distributed clusters for inference.

Table 3 provides a short summary of the models developed in the research as well as their training behavior.

**Table 3:** Quantitative summary of training performance across components.

| Model | Initial Loss | Final Loss | Epochs | Convergence Behavior |
|---|---|---|---|---|
| Autoencoder | 0.382 | 0.037 | 30 | Smooth convergence |
| Diffusion Model | 0.774 | 0.016 | 1,000 | Stable with minor stochastic fluctuations |
| MedChat | 2.45 | 0.23 | 4 | Stable convergence without overfitting |

# 4. Discussion

The developed medical chatbot, MedChat, is designed as a clinical decision support tool rather than a stand-alone diagnostic system. All interactions require supervision by qualified medical personnel, and no direct diagnostic or therapeutic functions are performed autonomously. Despite these constraints, the system demonstrates robust performance under realistic hardware conditions, validating the feasibility of offline deployment.

From a deployment perspective, MedChat is computationally accessible, as the base model can be run on consumer-grade GPUs equipped with at least 24 GB of VRAM, making it feasible for integration into local healthcare infrastructure without high-end computational resources. In addition, the Virtual Doctor avatar interface can run on 16 GB of VRAM or a reduced version where only audio is available can run without a graphic card and is even manageable by raspberry Pi 4b.

The current implementation of the avatar operates under a Markovian assumption, wherein each generated frame is conditioned solely on the preceding frame and the most recent audio segment. While this design choice significantly reduces computational complexity and enables efficient real-time generation, it also introduces limitations in temporal coherence. Specifically, the resulting motion between consecutive frames may appear subtle or slightly unnatural due to the absence of long-term temporal dependencies. Future improvements could involve extending the conditioning context through a 3D convolutional U-Net architecture [44], enabling the model to process and generate entire video sequences with awareness of multiple preceding and future frames. Such an approach would likely enhance temporal smoothness and realism in facial and gestural movements and thereby overcoming the uncanny valley. However, these benefits would come at the expense of increased computational requirements during both training and inference. Thus, the Markovian framework currently offers a pragmatic trade-off between performance efficiency and visual naturalness, serving as a foundation for subsequent advancements in temporally coherent avatar synthesis.

In terms of performance evaluation, the Virtual Doctor avatar, comprising the audio and image generation components, was qualitatively assessed using L1 loss to ensure fidelity to the expected latent space representations. However, such quantitative evaluation could not be directly applied to the MedChat dialogue model, due to the inherent subjectivity and open-ended nature of conversational interaction. Instead, the performance of MedChat will be systematically assessed through an upcoming study at the University Hospital Bochum, where its real-world applicability and usability will be evaluated under medical supervision.

Preliminary results, included in the appendix C, illustrate natural and coherent dialogue between the avatar and users, suggesting promising potential for interactive anamnesis. Nonetheless, several limitations must be acknowledged.

First, the training data for MedChat includes synthetically generated dialogues and relies heavily on the pretrained knowledge of the base LLaMA model [39], which may embed inherent biases. These biases could lead to unexpected behaviors or misinterpretations during the anamnesis process, particularly in complex or ambiguous clinical cases. It should always be used with care and medical supervision.

Second, the model's ability to generalize to rare symptoms or conditions is limited by the coverage and diversity of the synthetic dataset. As a result, MedChat may underperform when faced with less common medical presentations not adequately represented in the training data. Thus, it has to be improved in the future with corrected anamnesis results provided by physicians.

Third, the model's medical knowledge base is restricted to the information contained within its training corpus. This constraint implies that MedChat may lack awareness of emerging diseases, new clinical guidelines, or the latest evidence-based practices. This gap could compromise the accuracy and completeness of the anamnesis it conducts, highlighting the need for continued fine-tuning using large-scale, up-to-date medical corpora curated from verified sources.

In summary, MedChat represents a significant advancement in the development of AI-driven clinical support systems, combining natural language dialogue with a multimodal virtual avatar capable of generating realistic audio and visual outputs. This integration enhances the user experience by creating a more engaging and human-like interaction during the anamnesis process. While the current system has limitations related to training data coverage, potential biases, and the need for clinical validation, it already demonstrates the ability to conduct structured, medically relevant conversations in a natural and intuitive manner.

Importantly, MedChat is designed to run on accessible, consumer-grade hardware and follows a modular architecture, making it adaptable for a range of clinical environments. The forthcoming clinical evaluation and further fine-tuning on large-scale, domain-specific medical corpora are expected to improve its accuracy, generalizability, and safety. With continued development, MedChat, together with its Virtual Doctor avatar, has the potential to become a valuable tool in supporting healthcare professionals, streamlining patient interviews, and improving access to consistent, high-quality preliminary assessments. This work lays a strong foundation for the next generation of intelligent, multimodal healthcare assistants, capable of complementing human expertise and enriching patient care.

Future work will extend the temporal coherence of the avatar via 3D convolutional architectures and expand MedChat's conversational robustness through reinforcement learning with human feedback. The planned clinical evaluation at the University Hospital Bochum will further quantify usability, patient trust, and integration feasibility in real-world workflows.

# 5. Conclusion

In this paper, we present a fully AI-driven anamnesis pipeline designed to autonomously conduct patient interviews and generate concise, structured summaries to assist physicians by highlighting key medical information. The system, referred to as MedChat, integrates language models optimized for local execution, enabling deployment on consumer-grade graphics hardware. This ensures accessibility and practicality for clinical environments with limited computational infrastructure.

MedChat was developed using a combination of medical datasets and synthetically generated dialogues. However, it is important to note that the system has not yet undergone formal clinical validation. A prospective evaluation is planned at the University Hospital in Bochum, where the system's utility will be assessed from two perspectives: the extent to which it supports medical professionals in their diagnostic workflow, and the degree of acceptance and trust exhibited by patients toward an AI-guided anamnesis process.

Future development efforts will focus on integrating more advanced LLMs and enhancing the visual and interactive quality of the system's virtual avatar. A key ongoing challenge is maintaining a balance between system performance and practical feasibility. Many medical institutions operate under significant budget constraints and lack the resources required to deploy large-scale AI models. As such, our design prioritizes computational efficiency to offer a lightweight and cost-effective solution without compromising core functionality.

To promote transparency and foster further research in the field, we release both the synthetic training data and the developed neural network architectures as open-source software. This enables reproducibility of our results and provides a standardized benchmark for the development of future medical chatbot systems.

The system will be integrated into the Virtual Doctor project with testing and medical evaluation performed at the University Hospital Bochum in Germany. The results will be used to further improve on the framework and increase the benchmark dataset.

**Ethics Statement**
All experiments were performed using synthetic or publicly available datasets. No human or patient data were collected. All procedures comply with institutional and national ethical guidelines, including the EU General Data Protection Regulation (GDPR).

**Author Contributions**
D. H. served as the principal investigator and was responsible for overseeing the conceptualization and implementation of all components. J. B. R. developed the virtual avatar pipeline and supervised the development of the MedChat software. D. B. led the conceptualization and core development of the MedChat system. J. P. S. and A. C. contributed to the regulatory preparations necessary for future clinical testing of MedChat at the University Hospital Bochum, Germany.

**Funding**
Not available.

**Data Availability Statement**

Due to ethical and privacy considerations, the trained autoencoder and diffusion model weights are not openly released but can be made available upon reasonable research request and subject to appropriate data use agreements.

# References

[1] Jarrahi, Mohammad Hossein. (2018). Artificial Intelligence and the Future of Work: Human-AI Symbiosis in Organizational Decision Making. Business Horizons. 61. 10.1016/j.bushor.2018.03.007.

[2] Sarumi, Oluwafemi & Heider, Dominik. (2024). Large language models and their applications in bioinformatics. Computational and Structural Biotechnology Journal. 23. 10.1016/j.csbj.2024.09.031.

[3] Shi, Yi & Sun, Lin. (2024). How Generative AI Is Transforming Journalism: Development, Application and Ethics. Journalism and Media. 5. 582-594. 10.3390/journalmedia5020039.

[4] Bengio Y, Hinton G, Yao A, Song D, Abbeel P, Darrell T, Harari YN, Zhang YQ, Xue L, Shalev-Shwartz S, Hadfield G, Clune J, Maharaj T, Hutter F, Baydin AG, McIlraith S, Gao Q, Acharya A, Krueger D, Dragan A, Torr P, Russell S, Kahneman D, Brauner J, Mindermann S. Managing extreme AI risks amid rapid progress. Science. 2024 May 24;384(6698):842-845. doi: 10.1126/science.adn0117. Epub 2024 May 20. PMID: 38768279.

[5] Alberts IL, Mercolli L, Pyka T, Prenosil G, Shi K, Rominger A, Afshar-Oromieh A. Large language models (LLM) and ChatGPT: what will the impact on nuclear medicine be? Eur J Nucl Med Mol Imaging. 2023 May;50(6):1549-1552. doi: 10.1007/s00259-023-06172-w. Epub 2023 Mar 9. PMID: 36892666; PMCID: PMC9995718.

[6] Radford, A., Narasimhan, K., Salimans, T., & Sutskever, I. (2018). *Improving Language Understanding by Generative Pre-Training.* OpenAI Blog.

[7] Obradovich N, Khalsa SS, Khan W, Suh J, Perlis RH, Ajilore O, Paulus MP. Opportunities and Risks of Large Language Models in Psychiatry. NPP Digit Psychiatry Neurosci. 2024;2(1):8. doi: 10.1038/s44277-024-00010-z. Epub 2024 May 24. PMID: 39554888; PMCID: PMC11566298.

[8] Morley J, Machado CCV, Burr C, Cowls J, Joshi I, Taddeo M, Floridi L. The ethics of AI in health care: A mapping review. Soc Sci Med. 2020 Sep;260:113172. doi: 10.1016/j.socscimed.2020.113172. Epub 2020 Jul 15. PMID: 32702587.

[9] Tajabadi, Mohammad & Martin, Roman & Heider, Dominik. (2024). Privacy-Preserving Decentralized Learning Methods for Biomedical Applications. Computational and Structural Biotechnology Journal. 23. 10.1016/j.csbj.2024.08.024.

[10] Haenssle HA, Fink C, Schneiderbauer R, Toberer F, Buhl T, Blum A, Kalloo A, Hassen ABH, Thomas L, Enk A, Uhlmann L; Reader study level-I and level-II Groups; Alt C, Arenbergerova M, Bakos R, Baltzer A, Bertlich I, Blum A, Bokor-Billmann T, Bowling J, Braghiroli N, Braun R, Buder-Bakhaya K, Buhl T, Cabo H, Cabrijan L, Cevic N, Classen A, Deltgen D, Fink C, Georgieva I, Hakim-Meibodi LE, Hanner S, Hartmann F, Hartmann J, Haus G, Hoxha E, Karls R, Koga H, Kreusch J, Lallas A, Majenka P, Marghoob A, Massone C, Mekokishvili L, Mestel D, Meyer V,

Neuberger A, Nielsen K, Oliviero M, Pampena R, Paoli J, Pawlik E, Rao B, Rendon A, Russo T, Sadek A, Samhaber K, Schneiderbauer R, Schweizer A, Toberer F, Trennheuser L, Vlahova L, Wald A, Winkler J, Wölbing P, Zalaudek I. Man against machine: diagnostic performance of a deep learning convolutional neural network for dermoscopic melanoma recognition in comparison to 58 dermatologists. Ann Oncol. 2018 Aug 1;29(8):1836-1842. doi: 10.1093/annonc/mdy166. PMID: 29846502.

[11] Choi YJ, Park MJ, Ko Y, Soh MS, Kim HM, Kim CH, Lee E, Kim J. Artificial intelligence versus physicians on interpretation of printed ECG images: Diagnostic performance of ST-elevation myocardial infarction on electrocardiography. Int J Cardiol. 2022 Sep 15;363:6-10. doi: 10.1016/j.ijcard.2022.06.012. Epub 2022 Jun 9. PMID: 35691440.

[12] Gao Z, Xu Y, Sun C, Wang X, Guo Y, Qiu S, Ma K. A systematic review of asymptomatic infections with COVID-19. J Microbiol Immunol Infect. 2021 Feb;54(1):12-16. doi: 10.1016/j.jmii.2020.05.001. Epub 2020 May 15. PMID: 32425996; PMCID: PMC7227597.

[13] Tu, T., Schaekermann, M., Palepu, A. *et al.* Towards conversational diagnostic artificial intelligence. *Nature* (2025). https://doi.org/10.1038/s41586-025-08866-7

[14] Ran Xu, Wenqi Shi, Jonathan Wang, Jasmine Zhou, and Carl Yang. 2025. MedAssist: LLM-Empowered Medical Assistant for Assisting the Scrutinization and Comprehension of Electronic Health Records. In Companion Proceedings of the ACM on Web Conference 2025 (WWW '25). Association for Computing Machinery, New York, NY, USA, 2931–2934. https://doi.org/10.1145/3701716.3715186

[15] Lu, Meng & Ho, Brandon & Ren, Dennis & Wang, Xuan. (2024). TriageAgent: Towards Better Multi-Agents Collaborations for Large Language Model-Based Clinical Triage. 5747-5764. 10.18653/v1/2024.findings-emnlp.329.

[16] J V, Lavan & Sangeetha, Lakshmi. (2025). HealsHealthAI: Unveiling Personalized Healthcare Insights with Open Source Fine-Tuned LLM. 10.1002/9781394249312.ch4.

[17] Ruhland JB, Wichmann J, Degtyar D, Martin R, Fehse L, Klau JH, Papenbrock T, Sowa JP, Canbay A, Freisleben B, Leyer M, Heider D. The virtual doctor prescribing the future: Diagnostics with interactive clinical decision support. Comput Biol Med. 2025 Sep;196(Pt C):110968. doi: 10.1016/j.compbiomed.2025.110968. Epub 2025 Aug 20. PMID: 40839937.

[18] Yingbing Huang, Lily Jiaxin Wan, Hanchen Ye, Manvi Jha, Jinghua Wang, Yuhong Li, Xiaofan Zhang, and Deming Chen. 2024. Invited: New Solutions on LLM Acceleration, Optimization, and Application. In Proceedings of the 61st ACM/IEEE Design Automation Conference (DAC '24). Association for Computing Machinery, New York, NY, USA, Article 369, 1–4. https://doi.org/10.1145/3649329.3663517

[19] Winkler A, Kutschar P, Pitzer S, van der Zee-Neuen A, Kerner S, Osterbrink J, Krutter S. Avatar and virtual agent-assisted telecare for patients in their homes: A scoping review. J Telemed Telecare. 2025 Feb;31(2):207-221. doi: 10.1177/1357633X231174484. Epub 2023 Jun 7. PMID: 37287248.

[20] DeTore, Nicole & Balogun, Oyenike & Eberlin, Elizabeth & Dokholyan, Katherine & Rizzo, Albert & Holt, Daphne. (2024). An artificial intelligence-based virtual human avatar application to

assess the mental health of healthcare professionals: A validation study (Preprint). 10.2196/preprints.58255.

[21] Goodfellow, I. J., "Generative Adversarial Networks", arXiv:1406.2661, 2014. doi:10.48550/arXiv.1406.2661.

[22] Ho, J., Jain, A., and Abbeel, P., "Denoising Diffusion Probabilistic Models", Art. no. arXiv:2006.11239, 2020. doi:10.48550/arXiv.2006.11239.

[23] Deutch, G., Gal, R., Garibi, D., Patashnik, O., and Cohen-Or, D., "TurboEdit: Text-Based Image Editing Using Few-Step Diffusion Models", <i>arXiv e-prints</i>, Art. no. arXiv:2408.00735, 2024. doi:10.48550/arXiv.2408.00735.

[24] Salimans, T. and Ho, J., "Progressive Distillation for Fast Sampling of Diffusion Models", <i>arXiv e-prints</i>, Art. no. arXiv:2202.00512, 2022. doi:10.48550/arXiv.2202.00512.

[25] Baevski, A., Zhou, H., Mohamed, A., and Auli, M., "wav2vec 2.0: A Framework for Self-Supervised Learning of Speech Representations", <i>arXiv e-prints</i>, Art. no. arXiv:2006.11477, 2020. doi:10.48550/arXiv.2006.11477.

[26] J. Grosman, "Fine-tuned XLSR-53 large model for speech recognition in German," Hugging Face, 2021. [Online]. Available: https://huggingface.co/jonatasgrosman/wav2vec2-large-xlsr-53-german

[27] OHF-Voice, "piper1-gpl: Fast and local neural text-to-speech engine," Github. [Online]. Available: https://github.com/OHF-Voice/piper1-gpl

[28] Kong, J., Kim, J., and Bae, J., "HiFi-GAN: Generative Adversarial Networks for Efficient and High Fidelity Speech Synthesis", arXiv:2010.05646, 2020. doi:10.48550/arXiv.2010.05646.

[29] Kim, J., Kong, J., and Son, J., "Conditional Variational Autoencoder with Adversarial Learning for End-to-End Text-to-Speech", arXiv:2106.06103, 2021. doi:10.48550/arXiv.2106.06103.

[30] M. Hansen, "Thorsten medium checkpoint," Rhasspy, 2023. [Dataset]. Available: https://huggingface.co/datasets/rhasspy/piper-checkpoints/tree/main/de/de_DE/thorsten/medium

[31] M. Hansen, "Amy medium checkpoint," Rhasspy, 2023. [Dataset]. Available: https://huggingface.co/datasets/rhasspy/piper-checkpoints/tree/main/en/en_US/amy/medium

[32] Rombach, R., Blattmann, A., Lorenz, D., Esser, P., and Ommer, B., "High-Resolution Image Synthesis with Latent Diffusion Models", <i>arXiv e-prints</i>, Art. no. arXiv:2112.10752, 2021. doi:10.48550/arXiv.2112.10752.

[33] Shiry Ginosar, Kate Rakelly, Sarah Sachs, Brian Yin, Crystal Lee, Philipp Krähenbühl, Alexei A. Efros. A Century of Portraits: A Visual Historical Record of American High School Yearbooks, Extreme Imaging Workshop, International Conference on Computer Vision, ICCV 2015. and IEEE Transactions on Computational Imaging, September 2017.

[34] Jan Benedikt Ruhland, Iraj Masoudian, Dominik Heider, Enhancing deep neural network training through learnable adaptive normalization, Knowledge-Based Systems, Volume 326, 2025, 113968, ISSN 0950-7051, https://doi.org/10.1016/j.knosys.2025.113968.

[35] Nichol, A. and Dhariwal, P., "Improved Denoising Diffusion Probabilistic Models", <i>arXiv e-prints</i>, Art. no. arXiv:2102.09672, 2021. doi:10.48550/arXiv.2102.09672.

[36] Disease prediction based on symptoms. en. url: https://www.kaggle.com/datasets/pasindueranga/disease-prediction-based-on-symptoms (retrieved on 28.10.2024).

[37] Disease Prediction Using Machine Learning. url: https://www.kaggle.com/datasets/kaushil268/disease-prediction-using-machine-learning?resource=download (retrieved on 28. 10. 2024).

[38] Disease and Symptoms dataset. en. url: https://www.kaggle.com/datasets/choongqianzheng/disease-and-symptoms-dataset (retrieved on 28. 10. 2024).

[39] Abhimanyu Dubey u. a. The Llama 3 Herd of Models. arXiv:2407.21783. Aug. 2024. doi: 10.48550/arXiv.2407.21783. url: http://arxiv.org/abs/2407.21783

[40] Ilya Loshchilov und Frank Hutter. Decoupled Weight Decay Regularization. arXiv:1711.05101. Jan. 2019. doi: 10.48550/arXiv .1711.05101. url: http://arxiv.org/abs/1711.05101

[41] Edward J. Hu u. a. LoRA: Low-Rank Adaptation of Large Language Models. arXiv:2106.09685. Okt. 2021. doi: 10.48550/arXiv.2106.09685. url: http://arxiv.org/abs/2106.09685

[42] Yi Liu u. a. Prompt Injection attack against LLM-integrated Applications. arXiv:2306.05499. March 2024. doi: 10 . 48550 / arXiv . 2306 . 05499. url: http : //arxiv.org/abs/2306.05499

[43] meta-llama/Prompt-Guard-86M · Hugging Face. Sep. 2024. url: https://huggingface.co/meta-llama/Prompt-Guard-86M (visited on 29. 10. 2024).

[44] Ho, J., Salimans, T., Gritsenko, A., Chan, W., Norouzi, M., and Fleet, D. J., "Video Diffusion Models", arXiv:2204.03458, 2022. doi:10.48550/arXiv.2204.03458.

# Appendix A. Ablation Experiments

Our initial experiments were conducted using the selected dataset to design an autoencoder with a stable and well-constrained latent space, thereby improving training robustness during the subsequent diffusion modeling stage. To assess the impact of different latent space regularization strategies, we compared three autoencoder configurations based on the design described in table A1.

**Table A1:** Overview of the autoencoder architecture developed for this study. The model was implemented and tested across three configurations to investigate the influence of latent space regularization. In configuration 1, the tanh activation function at the encoder output was omitted. Configuration 2 integrated the reparameterization trick [32] to enforce a probabilistic latent distribution. Configuration 3, described in detail in this table, constitutes the final configuration adopted for all subsequent diffusion model experiments.

| Module | Layer / Operation | Configuration | Output Resolution | Notes |
|---|---|---|---|---|
| **Encoder** | Adaptive Normalization | AdaptiveNorm((1, in_channels, 1, 1)) | N×N | Input normalization |
| | Convolution | Conv2d(in_channels, 32, 3, 1, 1) | N×N | Initial feature extraction |
| | ResBlock | ResBlock(32, 32) | N×N | Residual feature refinement |
| | Downsampling | Conv2d(32, 32, 4, 2, 1) | N/2×N/2 | First spatial downsampling |
| | ResBlock | ResBlock(32, 64) | N/2×N/2 | Feature expansion |
| | Downsampling | Conv2d(64, 64, 4, 2, 1) | N/4×N/4 | Second spatial downsampling |
| | ResBlock | ResBlock(64, 128) | N/4×N4 | Deep feature encoding |

| | Downsampling | Conv2d(128, 128, 4, 2, 1) | N/8×N/8 | Latent compression |
| --- | --- | --- | --- | --- |
| | ResBlock | ResBlock(128, 128) | N/8×N/8 | High-level feature fusion |
| | Projection | Conv2d(128, latent_channels, 3, 1, 1) | N/8×N/8 | Latent embedding |
| | Activation | Tanh() | N/8×N/8 | Constrains latent space to [-1, 1] |
| **Decoder** | Adaptive Normalization | AdaptiveNorm((1, latent_channels, 1, 1)) | N/8×N/8 | Normalization of latent input |
| | Linear Projection | Conv2d(latent_channels, 128, 1) | N/8×N/8 | Expands latent space into feature domain |
| | ResBlock | ResBlock(128, 128) | N/8×N/8 | Residual learning in latent domain |
| | Upsampling | ConvTranspose2d(128, 64, 4, 2, 1) | N/4×N/4 | First spatial upsampling |
| | ResBlock | ResBlock(64, 64) | N/4×N/4 | Feature refinement |
| | Upsampling | ConvTranspose2d(64, 32, 4, 2, 1) | N/2×N/2 | Second spatial upsampling |
| | ResBlock | ResBlock(32, 32) | N/2×N/2 | Feature refinement |

| | Upsampling | ConvTranspose2d(32, 16, 4, 2, 1) | N×N | Final spatial upsampling |
|---|---|---|---|---|
| | ResBlock | ResBlock(16, 16) | N×N | Final feature smoothing |
| | Output Convolution | Conv2d(16, out_channels, 3, 1, 1) | N×N | Image reconstruction |
| | Activation | Tanh() | N×N | Constrains pixel values to [-1, 1] |

Our first approach was to establish a base line with an unconstrained latent space, i.e. where the latent variables were allowed to take arbitrary continuous values. The second approach was a variationally constrained latent space, where a normal prior distribution N(0,I) was enforced through the reparameterization trick and an additional Kullback-Leibler divergence loss term as described in [32]. The last approach contained a deterministically constrained latent space, where a hyperbolic tangent activation was applied to project latent variables into the range [-1,1]. An adaptive normalization layer [34] was incorporated in the decoder to re-establish consistent statistics during reconstruction.

The reparameterization trick introduces stochasticity into the latent representation while maintaining differentiability, enabling efficient gradient-based optimization. It can be expressed as:

(1) $$z=\mu+\sigma\times\epsilon, \quad \epsilon\sim N(0,I)$$

where $\mu$ and $\sigma$ represent the mean and standard deviation predicted by the encoder, and $\epsilon$ denotes a noise sample drawn from a standard normal distribution. To regularize the latent space toward a standard normal prior, we employed the Kullback-Leibler divergence loss, defined as:

(2) $$L_{KL} = -0.5 \sum_{i=1}^{d}(1+\log(\sigma_i^2)-\mu_i^2-\sigma_i^2)$$

where the sum is over the full dimensionality d of the latent space.

The reconstruction objective for all configurations was the L1 loss between the input and reconstructed image:

(3) $$L_{rec}=\|x-x\|^1_1$$

The autoencoder models were trained for 30 epochs using images of 256×256 resolution. To enhance the generalization capability of the network, a set of geometric data augmentation techniques was applied, including random rotations, translations, scalings, and shear transformations. All transformations employed zero padding to preserve image dimensions. The training objective combined an L1 reconstruction loss, as defined in equation (3), with a Kullback-Leibler divergence term serving as a regularization component for configuration 2. The regularization coefficient was varied between 0.000001and 0.01 to assess its impact on latent space stability. Optimization was performed using the AdamW algorithm [38], which provides adaptive moment estimation with decoupled weight decay to improve convergence behavior.

Figure A6 presents the evolution of the L1 reconstruction loss across all three autoencoder configurations. The relatively weaker reconstruction performance observed in configuration 2 at a regularization weight of 0.01 can be attributed to the increased stochasticity introduced by the KL divergence, which induces variability in the decoding process. When the regularization coefficient was reduced to 0.000001, the reconstruction quality improved. However, the latent space exhibited poor constraint, with activation magnitudes exceeding an order of magnitude of 10. Balancing the KL and L1 loss terms proved challenging, as excessive regularization reduced reconstruction fidelity, whereas insufficient regularization led to latent instability. Consequently, configuration 3 was selected as the final model, as it achieved a well-defined latent representation suitable for subsequent latent space noise augmentation, thereby enhancing the robustness of the downstream diffusion model.

Figure A7 provides a qualitative comparison between the original input images and their corresponding reconstructions for each autoencoder configuration, illustrating the inherent trade-off between reconstruction fidelity and latent space regularity. The results for the strongly KL-regularized autoencoder are omitted, as training exhibited strong KL divergence dominance. Consequently, although the latent space distribution closely approximated a normal prior, the reconstruction quality degraded substantially. This behavior, consistent with the quantitative results shown in figure A6, indicates that excessive KL regularization overly constrains the latent representation, thereby impairing the network's ability to accurately reconstruct fine-grained image details.

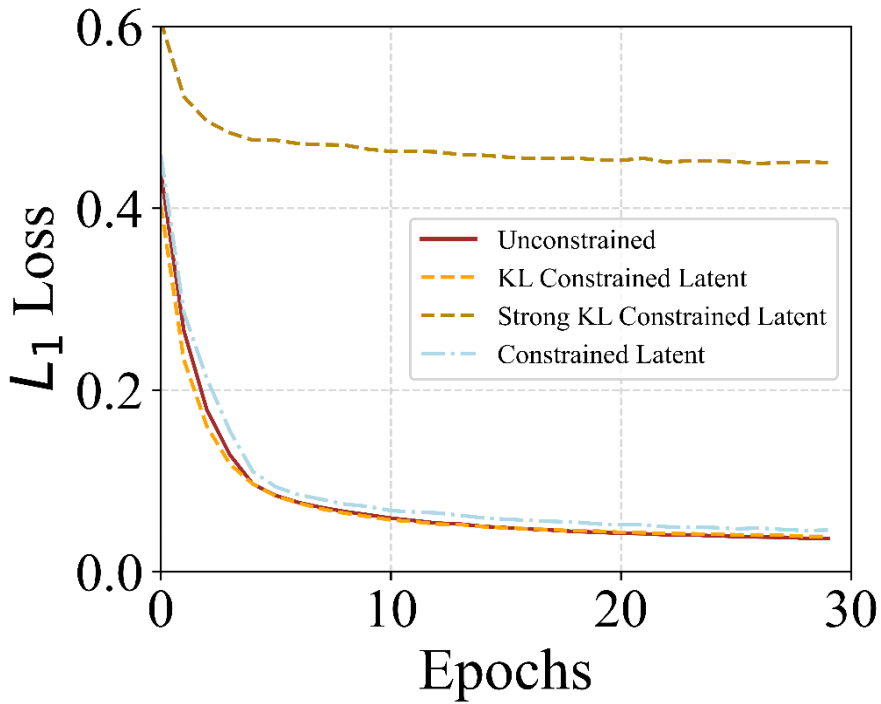

**Figure A6**: Training curves for the different autoencoder configurations. For configuration 2, which incorporates a KL divergence term, multiple regularization strengths were evaluated to assess the effect of relaxing the normal prior assumption. The curve labeled strong KL constraint corresponds to a regularization coefficient of 0.01, reflecting a strong enforcement of the prior, whereas the weak KL constraint (orange curve) employs a coefficient of 0.000001, allowing greater flexibility in the latent space representation.

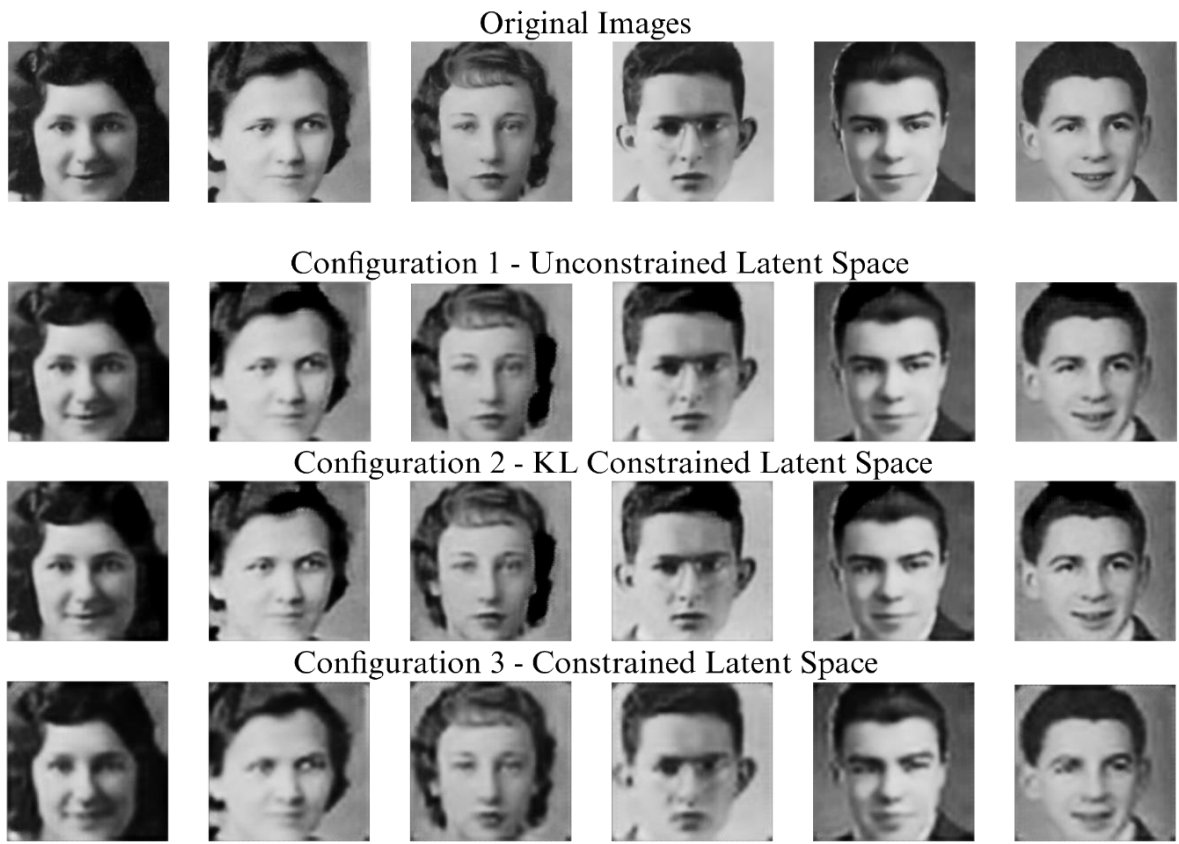

**Figure A7**: Examples of the original image and the reconstructed images using the autoencoder configurations.

To enhance robustness against residual noise introduced during the diffusion process, we augmented the latent space with Gaussian noise sampled from a normal distribution with a mean of 0 and a randomly varying variance. This strategy aimed to improve the model's generalization capability under stochastic perturbations during generation. The architecture of the subsequent U-Net model employed for diffusion-based synthesis is presented in table A2.

**Table A2:** Overview of the U-Net architecture employed in this study. The network incorporates adaptive normalization, cross-attention mechanisms, and time-conditioning via sinusoidal positional embeddings to enable temporally coherent generation. A Conditioning Encoder was introduced to integrate multimodal context from both image and mel-spectrogram inputs, enhancing semantic consistency during diffusion-based video synthesis. The detailed layer configuration, including downsampling, upsampling, and residual attention blocks, is summarized in the table.

| Module | Layer / Operation | Configuration / Parameters | Notes |
|---|---|---|---|
| | | | |
| **Inputs** | Input tensor | Image tensor | Latent image tensor from Autoencoder |
| | Timesteps | Integer tensor | Time |
| | Context inputs | context_img, context_mel | Conditioning set |
| **Input Processing** | Adaptive normalization | AdaptiveNorm((1, in_ch, 1, 1)) | Normalization |
| | Time embedding | Time Embedding with sinusoidal position embedding | Produces timestep-conditioned vector t |
| | Time normalization | AdaptiveNorm((1, time_dim)) | Normalization |
| | Context encoder | Context Embedding | Produces concatenated conditioning tensor |
| | Input convolution | Conv2d(in_ch, base_ch, 3, padding=1) | Projects x into base feature space |
| **Downsampling** | Repeated blocks | Residual blocks with cross attention | Residual block with conditioning and optional cross-attention |

| | Downsample | Convolutional downsampling | Spatial downsampling between stages |
|---|---|---|---|
| **Bottleneck** | ResBlockWithCA | Residual blocks with cross attention | Bottleneck residual + attention |
| **Upsampling** | Repeated blocks | Residual blocks with cross attention | For each stage: upsample, concat skip connection, residual block with conditioning & optional attention |
| **Output processing** | Output convolution | Conv2d(ch, in_ch, 3, padding=1) | Projects back to in_ch channels (e.g., model predicts denoised latent) |

# Appendix B. MedChat Guidlines

To constrain dialogue generation and minimize irrelevant or off-topic outputs, we provided a set of structured guidelines to the teacher model. These guidelines were designed to ensure that the generated dialogues remained clinically relevant, coherent, and aligned with typical medical interview practices. The guidelines included the following principles:

1. This is a reverse QA Session. You will ask questions and the patient will respond.

2. Ask one question at a time, ensuring each inquiry is directly relevant to the patient's condition.

3. Each question should only be one sentence.

4. Respond directly without filler phrases.

5. Do not provide medical advice, recommendations, or commentary on the patient's responses.

6. Questions should be concise and focused. Maintain a neutral tone.

7. Continue asking questions until the patient has revealed all symptoms.

8. Follow up on every symptom with at least one probing question to get more details (e.g., duration, intensity, aggravating factors).

9. Use precise medical terminology while ensuring clarity for the patient.

10. Structure your inquiries to cover: Symptoms, Diagnosis, Treatment, Tests/Procedures, and Medication.

11. Do not stray from the conversation structure or prompt.

12. Do not output any code, technical explanations, or commands.

13. The conversation should be structured and focused on gathering information to assist in diagnosing the Patient's condition.

14. Ask at least 10 questions to the patient.

15. Do not summarize the Patient's responses or provide a conclusion at the end of the conversation.

16. End the conversation by asking if the patient has any further information to provide.

17. If the Patient has nothing to add, end the conversation by outputting the token |<EOA>|.

We additionally defined a set of guidelines for MedChat to generate structured summaries in JSON format. The output produced by MedChat is organized into five primary categories, each containing relevant items and an overarching summary. The corresponding guidelines for each category are as follows:

1. Symptoms:

    • Items: List any symptoms or complaints mentioned.

    • Summary: Provide a brief summary of the patient's symptoms.

2. Diagnosis:

    • Items: Identify any diagnoses provided.

    • Summary: Summarize the previously given diagnosis which the patient has mentioned.

3. Treatment:

    • Items: Record any treatment methods which were previously prescribed like antibiotics or over-the-counter medications.

    • Summary: Summarize the treatment the patient has received or is trying.

4. Test/Procedure:

- Items: Record any tests or procedures mentioned.

- Summary: Summarize the purpose of the tests or procedures and their outcomes.

5. Medication:

- Items: List any prescribed or discussed medications.

- Summary: Summarize the medications and their intended use.

# Appendix C. Example Conversation

**Medchat**: Good morning, I'm Medchat, your AI healthcare expert. I'll do my best to help you today. Do you have any new complaints or symptoms that you'd like to discuss?

**Patient**: Yes, for several days I've had a high fever of around 39 °C.

**Medchat**: Have you experienced any chills or sweating along with this fever?

**Patient**: Yes, I'm sweating a lot and feeling fatigued.

**Medchat**: Have you noticed any changes in your appetite or any nausea or vomiting?

**Patient**: No.

**Medchat**: Have you had any recent travel, exposure to someone who's been sick, or contact with anything that might have caused an infection?

**Patient**: I don't think so. I do have a history of pneumonia and asthma.

**Medchat**: Have you noticed any shortness of breath or difficulty breathing, or are your asthma symptoms otherwise manageable?

**Patient**: Yes, I have shortness of breath and chest pain. I tried breathing exercises and had a chest X-ray.

**Medchat**: Did the chest X-ray show any abnormalities, and do you feel pain or tightness in your chest that worsens with deep breathing or coughing?

**Patient**: The X-ray was normal, but I do feel chest pain when coughing. I tried taking ibuprofen 600 mg, but it didn't help.

**Medchat**: Have you coughed up any mucus or blood, or experienced any pain or tenderness in your back or sides?

**Patient**: No mucus when coughing.

**Medchat**: Have you noticed any pain or pressure in your ears or head, or any recent changes in your hearing or vision?

**Patient**: No.

**Medchat**: Do you have any other information about your symptoms or medical history?

**Patient**: No, I don't.